\documentclass[letterpaper, 10 pt, conference]{ieeeconf}  

\IEEEoverridecommandlockouts                              
\makeatletter
\let\NAT@parse\undefined
\makeatother

\usepackage{algorithmic}
\usepackage{algorithm}
\usepackage[numbers,sort,comma]{natbib}
\usepackage{graphicx}
\usepackage{tabularx}
\usepackage{amsmath}
\usepackage{amssymb}
\usepackage{color}
\usepackage{graphicx,import}
\usepackage{booktabs}

\DeclareGraphicsExtensions{.pdf,.jpeg,.png}
\graphicspath{{figures/}}

\title{\LARGE \bf Collective Robot Reinforcement Learning with Distributed Asynchronous Guided Policy Search}

\author{Ali Yahya$^{1}$ \and Adrian Li$^{1}$ \and Mrinal Kalakrishnan$^{1}$ \and Yevgen Chebotar$^{2}$ \and Sergey Levine$^{3}$ 
\thanks{$^{1}$Ali Yahya, Adrian Li, and Mrinal Kalakrishnan are with X, Mountain View, CA 94043, USA.
{\tt\small \{alive,alhli,kalakris\}@x.team}}%
\thanks{$^{2}$Yevgen Chebotar is with the  Department of Computer Science, University of Southern California, Los Angeles, CA 90089, USA. This research was conducted during Yevgen's internship at X.}
\thanks{$^{3}$Sergey Levine is with Google Brain, Mountain View, CA 94043, USA.}
}

\begin{document}
\maketitle
\thispagestyle{empty}
\pagestyle{empty}

\begin{abstract}
In principle, reinforcement learning and policy search methods can enable robots to learn highly complex and general skills that may allow them to function amid the complexity and diversity of the real world. However, training a policy that generalizes well across a wide range of real-world conditions requires far greater quantity and diversity of experience than is practical to collect with a single robot. Fortunately, it is possible for multiple robots to share their experience with one another, and thereby, learn a policy collectively. In this work, we explore distributed and asynchronous policy learning as a means to achieve generalization and improved training times on challenging, real-world manipulation tasks. We propose a distributed and asynchronous version of Guided Policy Search and use it to demonstrate collective policy learning on a vision-based door opening task using four robots. We show that it achieves better generalization, utilization, and training times than the single robot alternative.
\end{abstract}

\section{Introduction}

Policy search techniques show promising ability to learn feedback control policies for robotic tasks with high-dimensional sensory inputs through trial and error \cite{TedrakeZS04,PetersMA10,TheodorouBS10,policysearch}. Most successful applications of policy search, however, rely on considerable manual engineering of suitable policy representations, perception pipelines, and low-level controllers to support the learned policy. Recently, deep reinforcement learning (RL) methods have been used to show that policies for complex tasks can be trained end-to-end, directly from raw sensory inputs (like images \cite{slmja-trpo-15,lhphe-ccdrl-16}) to actions. Such methods are difficult to apply to real-world robotic applications because of their high sample complexity. Methods based on Guided Policy Search (GPS) \cite{Levine:2016}, which convert the policy search problem into a supervised learning problem, with a local trajectory-centric RL algorithm acting as a teacher, reduce sample complexity and thereby help make said applications tractable. However, training such a policy to generalize well across a wide variety of real-world conditions requires far greater quantity and diversity of experience than is practical to collect with a single robot.

Fortunately, it is possible for multiple robots to share their experience with one another, and thereby, learn a policy collectively. In this work, we explore distributed and asynchronous policy learning (also known hereafter in this work as collective policy learning) as a means to achieve generalization and improved training times on challenging, real-world manipulation tasks. Collective policy learning presents a number of unique challenges. These challenges can be broadly categorized as utilization challenges and synchronization challenges. On one hand, we would like to maximize robot utilization \textemdash{} the fraction of time that the robots spend collecting experience for learning. On the other hand, each robot must allocate compute and bandwidth to process and communicate its experience to other robots, and the system as a whole needs to synchronize the assimilation of each robot's experience into the collective policy.

The main contribution of this work is a system for collective policy learning. We address the aforementioned utilization and synchronization challenges with a novel distributed and asynchronous variant of Guided Policy Search. In our system, multiple robots practice the task simultaneously, each on a distinct instance of the task, and jointly train a single policy whose parameters are maintained centrally by a parameter server. To maximize utilization, each robot continues to practice and optimize its own local policy while the single global policy is trained from a buffer of previously collected experience. For high-dimensional policies such as those based on neural networks, the increase in utilization that is conferred by asynchronous training is significant. Consequently, this approach dramatically brings down the total amount of time required to learn complex visuomotor policies using GPS, and makes this technique scalable to more realistic applications which require greater data diversity.

\begin{figure}
  \centering
  \includegraphics[trim=0cm 0 0cm 0, clip=true,width=0.588\columnwidth]{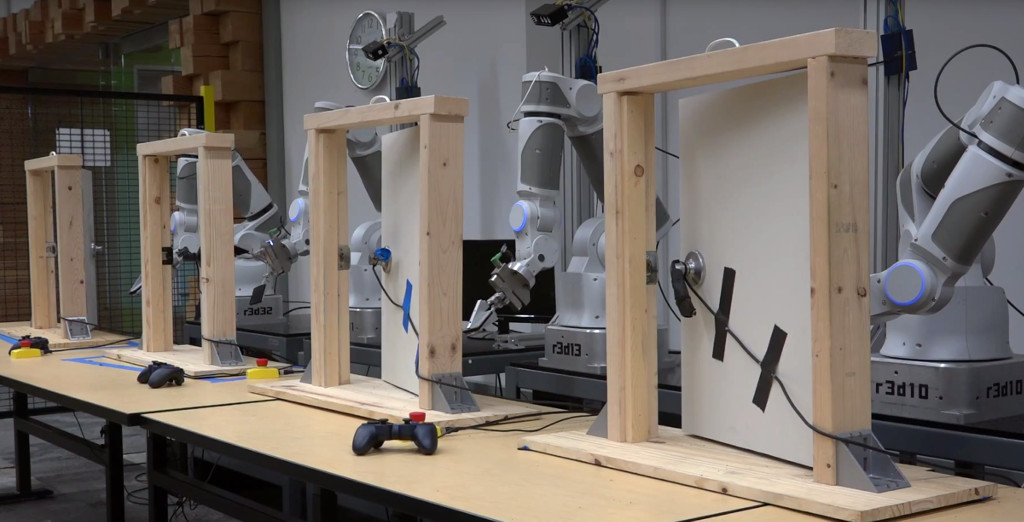}
  \includegraphics[trim=0cm 0 0cm 0, clip=true,width=0.393\columnwidth]{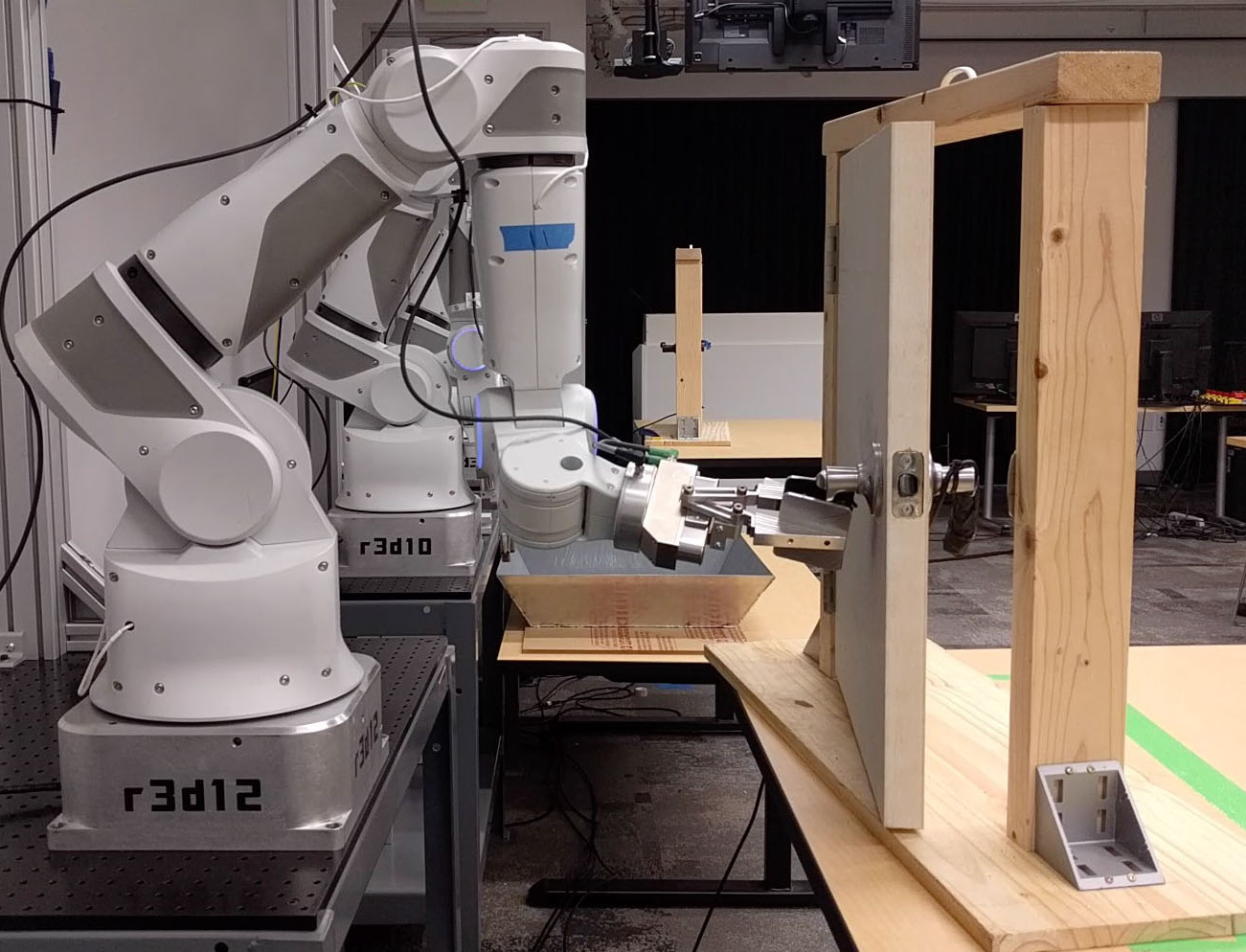}
  \caption{Multiple robots collaborating to learn the door opening skill. Our system allows the robots to operate continuously to collect a large amount of diverse experience, while the policy is simultaneously trained with a replay buffer of the latest trajectory samples.}
  \label{fig:teaser}
\end{figure}

We evaluate our approach in simulation and on a real-world door opening task (shown in Figure~\ref{fig:teaser}), where both the pose and the appearance of the door vary across task instances. We show that our system achieves better generalization, utilization, and training times than the single robot alternative.

\section{Related Work}

Robotic motor skill learning has shown considerable promise for enabling robots to autonomously learn complex motion skills \cite{TedrakeZS04,PetersMA10,TheodorouBS10,policysearch}. However, most successes in robotic motor skill learning have involved significant manual design of representations in order to enable policies to generalize effectively. For example, the well-known dynamic movement primitive representation \cite{dmp} has been widely used to generalize learned skills by adapting the goal state, but it inherently restricts the learning process to trajectory-centric behaviors. 

Enabling robotic learning with more expressive policy classes that can represent more complex strategies has the potential of eliminating the need for the manual design of representations. Recent years have seen improvement in the generalizability of passive perception systems, in domains such as computer vision, natural language processing, and speech recognition through the use of deep learning techniques \cite{deep_learning}. These methods combine deep neural networks with large datasets to achieve remarkable results on a diverse range of real-world tasks. However, the requirement of large labeled datasets has limited the application of such methods to robotic learning problems. While several works have extended deep learning methods to simulated \cite{slmja-trpo-15,lhphe-ccdrl-16} and real-world \cite{LampeRiedmiller2013,Levine:2016} robotic tasks, the kind of generalization exhibited by deep learning in passive perception domains has not yet been demonstrated for robotic skill learning. This may be due to the fact that robotic learning experiments tend to use relatively small amounts of data in constrained domains, with a few hours of experience collected from a single robot in each experiment.

A central motivation behind our work is the ability to apply deep learning to robotic manipulation by making it feasible to collect large amounts of on-policy experience with real physical platforms. While this may seem impractical for small-scale laboratory experiments, it becomes much more realistic when we consider a possible future where robots are deployed in the real-world to perform a wide variety of skills. The challenges of asynchrony, utilization, and parallelism, which we aim to address in this work, are central for such real-world deployments. The ability of robotic systems to learn more quickly and effectively by pooling their collective experience has long been recognized in the domain of cloud robotics, where it is typically referred to as collective robotic learning \cite{ikkhi-prrbr-00,k-cehr-10,kmckg-cbrgg-13,kpag-srcra-15}. Our work therefore represents a step toward more practical and powerful collective learning with distributed, asynchronous data collection.

Distributed systems have long been an important subject in deep learning \cite{parallel_sgd}. While distributed asynchronous architectures have previously been used to optimize controllers for simulated characters \cite{mordatch2015}, our work is, to the best of our knowledge, the first to experimentally explore distributed asynchronous training of deep neural network policies for real-world robotic control. In our work, we parallelize both data collection and neural network policy training across multiple machines.

\section{Preliminaries on Guided Policy Search}

In this section, we define the problem formulation and briefly summarize theb guided policy search (GPS) algorithm, specifically pointing out computational bottlenecks that can be alleviated through asynchrony and parallelism. A more complete description of the theoretical underpinnings of the method can be found in prior work~\cite{Levine:2016}. The goal of policy search methods is to optimize the parameters $\theta$ of a policy $\pi_\theta(\mathbf{u}_t | \mathbf{x}_t)$, which defines a probability distribution over robot actions $\mathbf{u}_t$ conditioned on the system state $\mathbf{x}_t$ at each time step $t$ of a task execution. Let $\tau = (\mathbf{x}_1, \mathbf{u}_1, \dots,  \mathbf{x}_T, \mathbf{u}_T)$ be a trajectory of states and actions. Given a task cost function $l(\mathbf{x}_t, \mathbf{u}_t)$, we define the trajectory cost $l(\tau)=\sum_{t=1}^T l(\mathbf{x}_t, \mathbf{u}_t)$. Policy optimization is performed with respect to the expected cost of the policy:
\[
J(\theta) = E_{\pi_\theta}\left[l(\tau)\right] = \int l(\tau) p_{\pi_\theta} (\tau) d\tau,
\]
where $p_{\pi_\theta} (\tau)$ is the policy trajectory distribution given the system dynamics $p\left(\mathbf{x}_{t+1} | \mathbf{x}_{t}, \mathbf{u}_{t}\right)$:
\[
p_{\pi_\theta} (\tau) = p(\mathbf{x}_1) \prod_{t=1}^{T} p\left(\mathbf{x}_{t+1} | \mathbf{x}_{t}, \mathbf{u}_{t}\right) \pi_\theta(\mathbf{u}_t | \mathbf{x}_t).
\]
Most standard policy search methods aim to directly optimize this objective, for example by estimating the gradient $\nabla_\theta J(\theta)$. However, this kind of direct model-free method can quickly become intractable for very high-dimensional policies, such as the large neural network policies considered in this work \cite{policysearch}. An alternative approach is to train the deep neural network with supervised learning, using a simpler local policy optimization method to produce supervision. To that end, guided policy search introduces a two-step approach for learning high-dimensional policies by combining the benefits of simple, efficient trajectory-centric RL and supervised learning of high-dimensional, nonlinear policies. Instead of directly learning the policy parameters with reinforcement learning, a trajectory-centric algorithm is first used to learn simple local controllers $p_i(\mathbf{u}_t | \mathbf{x}_t)$ for trajectories with various initial conditions, which might correspond, for instance, to different poses of a door for a door opening task. We refer to these controllers as \textit{local policies}. In this work, we employ time-varying linear-Gaussian controllers of the form $p_i(\mathbf{u}_t | \mathbf{x}_t) = \mathcal{N}(\mathbf{K}_{t} \mathbf{x}_t + \mathbf{k}_{t}, \mathbf{C}_{t})$ to represent these local policies, following prior work~\cite{Levine:2016}.

After optimizing local policies, the controls from these policies are used to create a training set for learning a complex high-dimensional \textit{global policy} in a supervised manner. Hence, the final global policy generalizes to the initial conditions of multiple local policies and can contain thousands of parameters, which can be efficiently learned with supervised learning. Furthermore, while trajectory optimization might require the full state $\mathbf{x}_t$ of the system to be known, it is possible to only use the observations $\mathbf{o}_t$ of the full state for training a global policy $\pi_\theta(\mathbf{u}_t | \mathbf{o}_t)$. This allows the global policy to predict actions from raw observations at test time~\cite{Levine:2016}.

In this work, we will examine a general asynchronous framework for guided policy search algorithms, and we will show how this framework can be instantiated to extend two prior guided policy search methods: BADMM-based guided policy search~\cite{Levine:2016} and mirror descent guided policy search (MDGPS)~\cite{MontgomeryL16}. Both algorithms share the same overall structure, with alternating optimization of the local policies via trajectory-centric RL, which in the case of our system is either a model-based algorithm based on LQR~\cite{LevineA14} or a model-free algorithm based on PI$^2$~\cite{TheodorouBS10}, and optimization of the global policy via supervised learning through stochastic gradient descent (SGD). The adaptation of PI$^2$ to guided policy search is described in detail in a companion paper~\cite{chebotar-icra2017}. The difference between the two methods is the mechanism that is used to keep the local policies close to the global policy. This is extremely important, since in general not all local policies can be reproduced effectively by a single global policy.

\paragraph{BADMM-based GPS} In BADMM-based guided policy search~\cite{Levine:2016}, the alternating optimization is formalized as a constrained optimization of the form
\begin{align*}
\! \min_{\theta,p_1,\dots,p_N} \!\! \sum_{i=1}^N E_{\tau \sim p_i}[l(\tau)] \text{ s.t. } p_i(\mathbf{u}_t | \mathbf{x}_t) \!=\! \pi_\theta(\mathbf{u}_t | \mathbf{x}_t) \, \forall \, \mathbf{x}_t,\mathbf{u}_t,i.
\end{align*}
This is equivalent in expectation to optimizing $J(\theta)$, since $\sum_{i=1}^N E_{\tau \sim p_i}[l(\tau)] = \sum_{i=1}^N E_{\tau \sim \pi_\theta}[l(\tau)]$ when the constraint is satisfied, and $\sum_{i=1}^N E_{\tau \sim \pi_\theta}[l(\tau)] \approx E_{\mathbf{x}_1 \sim p(\mathbf{x}_1), \tau \sim \pi_\theta}[l(\tau)]$ when the initial states $\mathbf{x}_1^i$ are sampled from $p(\mathbf{x}_1)$. The constrained optimization is then solved using the Bregman ADMM algorithm~\cite{WangB14}, which augments the objective for both the local and global policies with Lagrange multipliers that keep them similar in terms of KL-divergence. These terms are denoted $\phi_i(\tau,\theta,\tau)$ and $\phi_\theta(p_i,\theta,\tau)$ for the local and global policies, respectively, so that the global policy is optimized with respect to the objective
\begin{equation}
\min_\theta \! \sum_{i=1}^N E_{\tau \sim p_i} \!\!\left[\sum_{t=1}^T D_\text{KL}(\pi_\theta(\mathbf{u}_t|\mathbf{x}_t) \| p_i(\mathbf{u}_t|\mathbf{x}_t)) \!+\! \phi_\theta(p_i,\theta,\tau)\right]\!\!,\label{eqn:badmmsup}
\end{equation}
and the local policies are optimized with respect to
\begin{equation}
\min_{p_i} E_{\tau \sim p_i(\tau)}[l(\tau)] \text{ s.t. } D_\text{KL}(p_i(\tau) \| \bar{p}_i(\tau)) < \epsilon, \label{eqn:badmmtraj}
\end{equation}
where $\bar{p}_i$ is the local policy at the previous iteration. The constraint ensures that the local policies only change by a small amount at each iteration, to prevent divergence of the trajectory-centric RL algorithm, analogously to other recent RL methods \cite{PetersMA10,daniel2012hierarchical}. The derivations of $\phi_i(\tau,\theta,\mathbf{x}_t)$ and $\phi_\theta(p_i,\theta,\mathbf{x}_t)$ are provided in prior work~\cite{Levine:2016}.

\paragraph{MDGPS} In MDGPS~\cite{MontgomeryL16}, the local policies are optimized with respect to
\begin{equation}
\min_{p_i} E_{\tau \sim p_i(\tau)}[l(\tau)] \text{ s.t. } D_\text{KL}(p_i(\tau) \| \pi_\theta(\tau)) < \epsilon, \label{eqn:mdgpstraj}
\end{equation}
where the constraint directly limits the deviation of the local policies from the global policies. This can be interpreted as the generalized gradient step in mirror descent, which improves the policy with respect to the objective. The supervised learning step simply minimizes the deviation from the local policies, without any additional augmentation terms:
\begin{equation}
\min_\theta \! \sum_{i=1}^N E_{\tau \sim p_i} \!\!\left[\sum_{t=1}^T D_\text{KL}(\pi_\theta(\mathbf{u}_t|\mathbf{x}_t) \| p_i(\mathbf{u}_t|\mathbf{x}_t))\right]. \label{eqn:mdgpssup}
\end{equation}
This can be interpreted as the projection step in mirror descent, such that the overall algorithm optimizes the global policy subject to the constraint that the policy should lie within the manifold defined by the policy class, which in our case corresponds to neural networks.

\begin{algorithm}[t]
	\caption{Standard synchronous guided policy search \label{alg:gps}}
	\begin{algorithmic}[1]
		\FOR{iteration $k \in \{1, \dots, K\}$}
		\STATE Generate sample trajectories starting from each $\mathbf{x}_1^i$ by executing $p_i(\mathbf{u}_t|\mathbf{x}_t)$ or $\pi_\theta(\mathbf{u}_t|\mathbf{x}_t)$ on the robot.
		\STATE Use samples to optimize each of the local policies $p_i(\mathbf{u}_t|\mathbf{x}_t)$ from each $\mathbf{x}_1^i$ with respect to Equation~(\ref{eqn:badmmtraj}) or (\ref{eqn:mdgpstraj}), using either LQR or PI$^2$.
		\STATE Optimize the global policy $\pi_\theta(\mathbf{u}_t|\mathbf{x}_t)$ according to Equation~(\ref{eqn:badmmsup}) or (\ref{eqn:mdgpssup}) with SGD.
		\ENDFOR
	\end{algorithmic}
\end{algorithm}

Both GPS algorithms are summarized in Algorithm~\ref{alg:gps}, and consist of two alternating phases: optimization of the local policies with trajectory-centric RL, and optimization of the global policy with supervised learning. Several different trajectory-centric RL algorithms may be used, and we summarize the ones used in our experiments below.

\subsection{Local Policy Optimization}

The GPS framework is generic with respect to the choice of local policy optimizer. In this work, we consider two possible methods for local policy optimization:

\paragraph{LQR with local models} To take a model-based approach to optimization of time-varying linear-Gaussian local policies, we can observe that, under time-varying linear-Gaussian dynamics, the local policies can be optimized analytically using the LQR method, or the iterative LQR method in the case of non-quadratic costs \cite{LiT04}. However, this approach requires a linearization of the system dynamics, which are generally not known for complex robotic manipulation tasks. As described in prior work \cite{LevineA14}, we can still use LQR if we fit a time-varying linear-Gaussian model to the samples using linear regression. In this approach, the samples generated on line 1 of Algorithm~\ref{alg:gps} are used to fit a time-varying linear-Gaussian dynamics model of the form $p(\mathbf{x}_{t+1}|\mathbf{x}_t,\mathbf{u}_t) = \mathcal{N}(\mathbf{F}_{\mathbf{x},t}\mathbf{x}_t + \mathbf{F}_{\mathbf{u},t}\mathbf{u}_t + \mathbf{f}_t, \mathbf{N}_t)$. As suggested in prior work, we can use a Gaussian mixture model (GMM) prior to reduce the sample complexity of this linear regression fit, and we can accommodate the constraints in Equation~(\ref{eqn:badmmtraj}) or (\ref{eqn:mdgpstraj}) with a simple modification that uses LQR within a dual gradient descent loop \cite{LevineA14}.

\paragraph{PI$^2$}
Policy Improvement with Path Integrals (PI$^2$) is a model-free policy search method based on the principles of stochastic optimal control \cite{TheodorouBS10}. It does not require fitting of linear dynamics and can be applied to tasks with highly discontinuous dynamics or non-differentiable costs. In this work, we employ PI$^2$ to learn feedforward commands $\mathbf{k}_t$ of time-varying linear-Gaussian controllers as described in \cite{chebotar-icra2017}. 

The controls at each time step are updated according to the soft-max probabilities $P_{i,t}$ based on their cost-to-go $S_{i,t}$:
\[
S_{i,t} = S(\tau_{i,t}) = \sum^T_{j=t} l(\mathbf{x}_{i,j}, \mathbf{u}_{i,j}), \,\,\, P_{i,t} = \frac{ e^{-\frac{1}{\eta} S_{i,t}}}{\sum_{i=1}^N e^{-\frac{1}{\eta} S_{i,t}}},
\]
where $l(\mathbf{x}_{i,j}, \mathbf{u}_{i,j})$ is the cost of sample $i$ at time $j$. In this way, trajectories with lower costs become more probable after the policy update. For learning feedforward commands, the policy update corresponds to a weighted maximum likelihood estimation of the new mean $\mathbf{k}_t$ and the noise covariance $\mathbf{C}_t$. In this work, we  use relative entropy optimization  \cite{PetersMA10} to determine the temperature $\eta$ at each time step independently, based on a KL-divergence constraint between policy updates.

Both the LQR model-based method and the PI$^2$ model-free algorithm require samples in order to improve the local policies. In the BADMM variant of GPS, these samples are always generated from the corresponding local policies. However, in the case of MDGPS, the samples can in fact be generated directly by the global policy, with the local policies only existing temporarily within each iteration for the purpose of policy improvement. In this case, new initial states can be sampled at each iteration of the algorithm, with new local policies instantiated for each one~\cite{MontgomeryL16}. We make use of this capability in our experiments to train the global policy on a wider range of initial states in order to improve generalization.

\section{Asynchronous Distributed Guided Policy Search}

\begin{figure}
\centering
\includegraphics[width=\columnwidth]{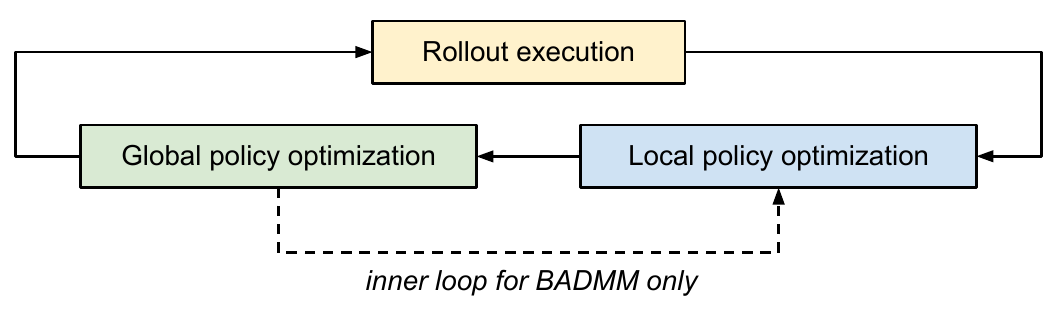}
\caption{Diagram of the training loop for synchronous GPS with a single replica. Rollout execution corresponds to line 2 in Algorithm~\ref{alg:gps}, local policy optimization to line 3, and global policy optimization to line 4. In BADMM-based GPS, the algorithm additionally alternates between local and global policy optimization multiple times before executing new rollouts. This sequential version of the algorithm requires training to pause while performing rollouts, and vice versa.}
\label{fig:gps}
\end{figure}

In synchronous GPS, rollout execution and policy optimization occur sequentially (see Figure~\ref{fig:gps}). This training regime presents two challenges: (1) there is considerable downtime for the robot while the policies are being optimized, and (2) there are synchronization issues when extending the global policy optimization to use data collected across multiple robots.

\begin{figure}
\centering
\includegraphics[width=0.65\columnwidth]{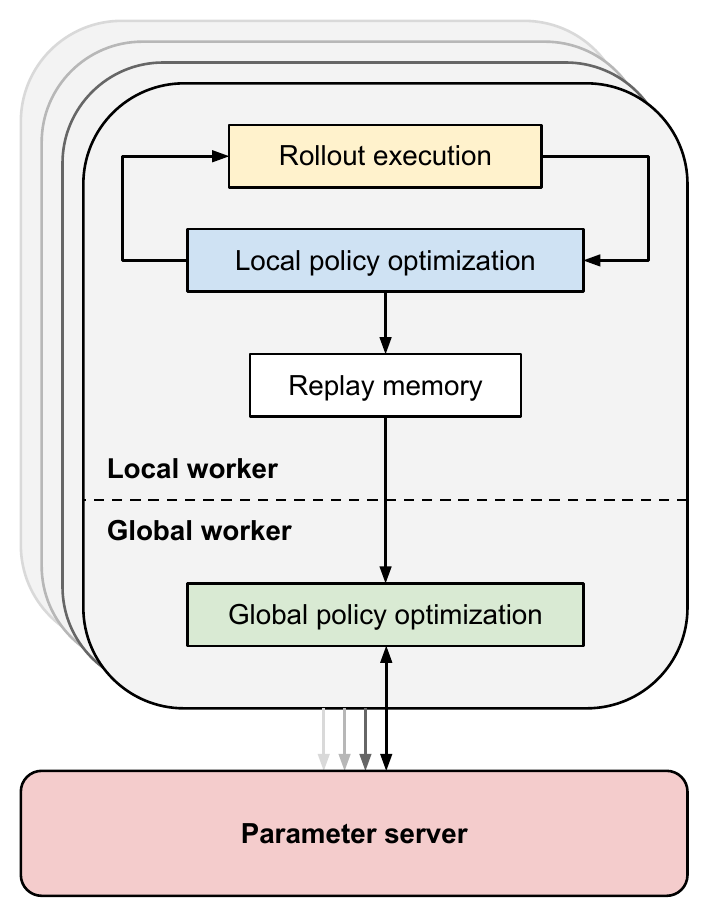}
\caption{The training loop for ADGPS with multiple replicas. Rollout execution and global policy optimization are decoupled via the replay memory. Multiple robots concurrently collect data and asynchronously update the parameter server, allowing maximal utilization of both computational and robot resources, as well as parallelization across multiple robots and servers.}
\label{fig:adgps}
\end{figure}

To overcome these challenges, we propose a modification to GPS which is both asynchronous and distributed (see Figure~\ref{fig:adgps}). In our asynchronous distributed GPS method (ADGPS), the algorithm is decoupled into global and local worker threads. The global workers are responsible for continuously optimizing the global policy using a buffer of experience data, which we call the replay memory. The local workers execute the current controllers on their respective robots, adding the collected data to the replay memory. The local workers are also responsible for updating the local policies. Note, however, that updating the local policies is very quick when compared to global policy updates, since the local policy update requires either a small number of LQR backward passes, or simply a weighted average of the sample controls, if using the PI$^2$ method. This operation can be completed in just a few seconds, while global policy training requires stochastic gradient descent (SGD) optimization of a deep neural network, and can take hours.

The local and global worker threads communicate through the replay memory, which stores the rollouts and optimized trajectories from each local worker. Since the rollouts in this memory are not guaranteed to come from the latest policy, they are reweighted at every iteration using importance sampling. The global workers asynchronously read from the replay memory and apply updates to the global policy. By decoupling the local and global work, the robots can now continuously collect data by executing rollouts, while the global policy is optimized in the background. This system also makes it easy to add multiple robots into the training process, by adding additional local workers for every robot.

The global policy itself can be represented with any function approximator, but in our work, as in prior GPS methods, we use a deep neural network representation trained with stochastic gradient descent (SGD), which can itself be trained in a distributed manner. The global policy is stored on a parameter server~\cite{dean2012large}, allowing multiple robots to concurrently collect data while multiple machines concurrently apply updates to the same global policy. By utilizing more robots, we are able to achieve much greater data diversity than would otherwise be realized with only a single robot, and by using multiple global worker threads, we can accelerate global policy training.

The replay memory may be either centralized or distributed. In our implementation of this system, each physical machine connected to each physical robot maintains its own replay memory. This is particularly convenient if we also run a single global worker thread on each physical machine, since it removes the need to transmit the high-bandwidth rollout data between machines during training. Instead, the machines only need to communicate model parameter updates to the centralized parameter server, which are typically much smaller than images or high-frequency joint angle and torque trajectories. In this case, the only centralized element of this system is the parameter server. Furthermore, since mini-batch gradient descent assumes uncorrelated examples within each batch, we found that distributed training actually improved stability when aggregating gradients across multiple robots and machines~\cite{chen-iclr2016}.

\begin{algorithm}[t]
	\caption{Asynchronous distributed guided policy search (local worker) \label{alg:adgps-local}}
	\begin{algorithmic}[1]
		\FOR{iteration $k \in \{1, \dots, K\}$}
		\STATE Generate sample trajectories starting from each $\mathbf{x}_1^i$ assigned to this worker, by executing either $p_i(\mathbf{u}_t|\mathbf{x}_t)$ or $\pi_\theta(\mathbf{u}_t|\mathbf{x}_t)$ on the robot.
		\STATE Use samples to optimize each of the local policies $p_i(\mathbf{u}_t|\mathbf{x}_t)$ from each $\mathbf{x}_1^i$ with respect to Equation~(\ref{eqn:badmmtraj}) or (\ref{eqn:mdgpstraj}), using either LQR or PI$^2$.
		\STATE Append optimized trajectories $p_i(\mathbf{u}_t|\mathbf{x}_t)$ to replay memory $\mathcal{D}$.
		\ENDFOR
	\end{algorithmic}
\end{algorithm}

\begin{algorithm}[t]
	\caption{Asynchronous distributed guided policy search (global worker) \label{alg:adgps-global}}
	\begin{algorithmic}[1]
		\FOR{step $n \in \{1, \dots, N\}$}
		\STATE Randomly sample a mini-batch $\{\mathbf{x}_t^j\}$ from the replay memory $\mathcal{D}$, with corresponding labels obtained from the corresponding local policies $p_i(\mathbf{u}_t|\mathbf{x}_t^j)$, where $i$ is the instance from which sample $j$ was obtained.
		\STATE Optimize the global policy $\pi_\theta(\mathbf{u}_t|\mathbf{x}_t)$ on this mini-batch for one step of SGD with respect to Equation~(\ref{eqn:badmmsup}) or (\ref{eqn:mdgpssup}).
		\ENDFOR
	\end{algorithmic}
\end{algorithm}

This entire system, which we call asynchronous distributed guided policy search (ADGPS), was implemented in the distributed machine learning framework TensorFlow \cite{tensorflow2015-whitepaper}, and is summarized in Algorithms~\ref{alg:adgps-local} and \ref{alg:adgps-global}. In our implementation, rollout execution and local policy optimization are still performed sequentially on the local worker as the optimization is a relatively cheap step; however, this is not strictly necessary and both steps could also be performed asynchronously. It is also possible to instantiate this system with varying numbers of global and local workers, or even a single centralized global worker. However, as discussed above, associating a single global worker with each local worker allows us to avoid transmitting the rollout data between machines, leading to a particularly efficient and convenient implementation.

\section{Experimental Evaluation}

Our experimental evaluation aims to answer two questions about our proposed asynchronous learning system: (1) does distributed asynchronous learning accelerate the training of complex, nonlinear neural network policies, and (2) does training across an ensemble of multiple robots improve the generalization of the resulting policies. The answers to these questions are not trivial: although it may seem natural that parallelizing experience collection should accelerate learning, it is not at all clear whether the additional bias introduced by asynchronous training would not outweigh the benefit of greater dataset size, nor that the amount of data is indeed the limiting factor.

\subsection{Simulated Evaluation}
\label{sec:simulated_evaluation}

In simulation, we can systematically vary the number of robots to evaluate how training times scale with worker count, as well as study the effect of asynchronous training. We simulated multiple 7-DoF arms with parallelized simulators that each run in real time, in order to mimic rollout execution times that would be observed on real robots. The arms are controlled with torque control in order to perform a simple Cartesian reaching task that requires placing the end-effector at a commanded position. The robot state vector consists of joint angles and velocities, as well as its end-effector pose and velocity. We use a 9-DoF parameterization of pose, containing the positions of three points rigidly attached to the robot end-effector represented in the base frame. The Cartesian goal pose uses the same representation, and is fed to the global policy along with the robot state. The global policy must be able to place the end-effector at a variety of different target positions, with each instance of the task corresponding to a different target. We train the policy using 8 instances of the task, using 4 additional instances as a validation set for hyperparameter tuning, and finally test the global policy on 4 held-out instances. These experiments use the LQR variant of BADMM-based GPS.

\begin{figure}
  \centering
  \includegraphics[trim=0pt -7pt 0cm 0, clip=true,width=0.49\columnwidth]{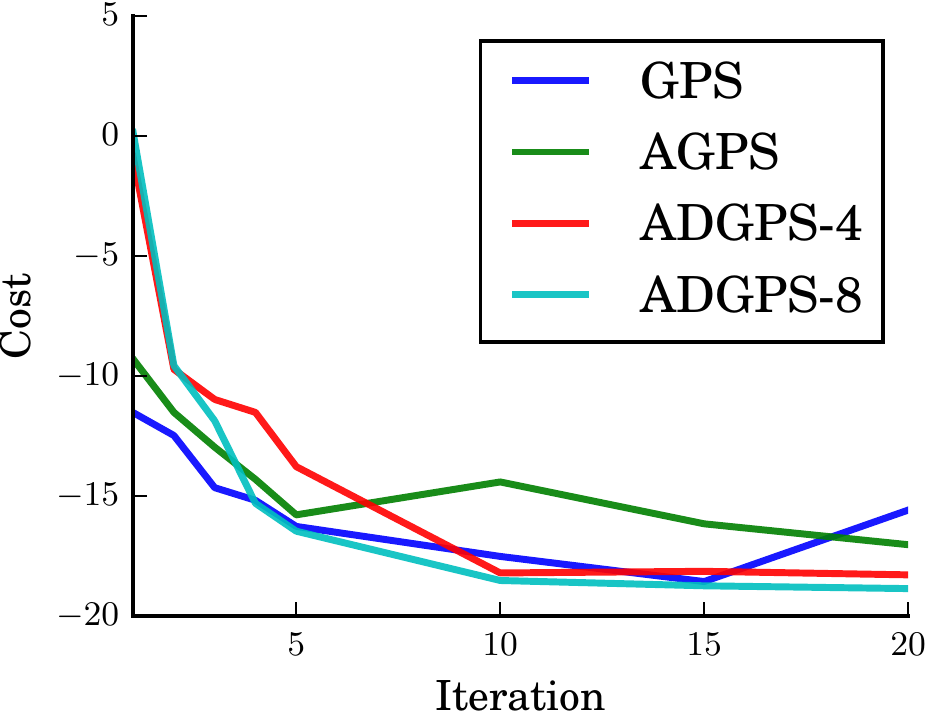}%
  \includegraphics[trim=-7pt 0 0cm 0, clip=true,width=0.49\columnwidth]{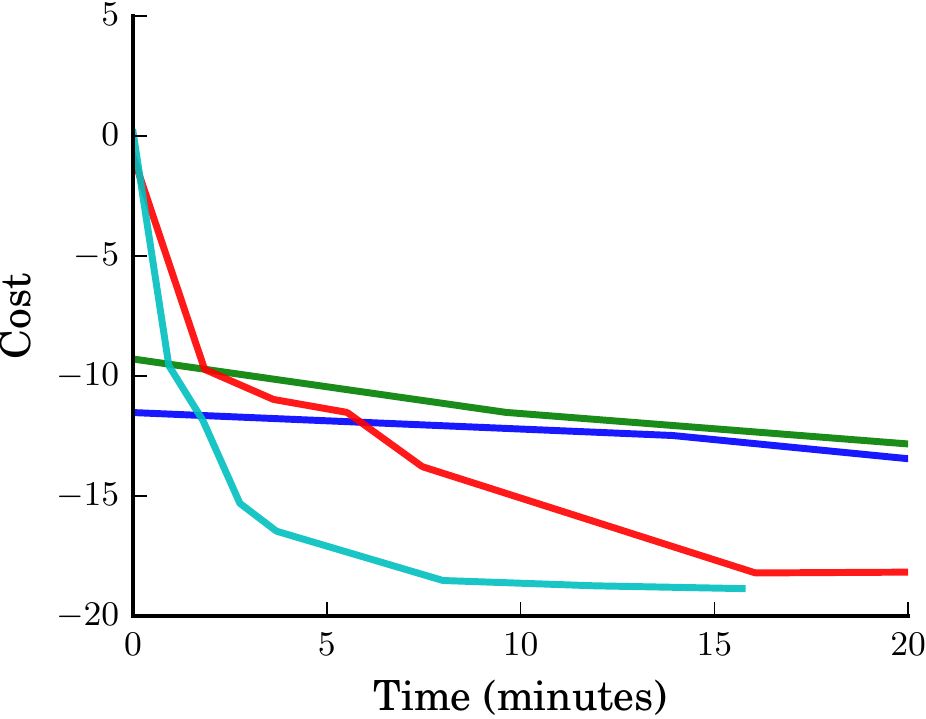}
  \vspace{-5pt}
  \caption{Average costs of the 4 test instances used in the simulated reaching task, over number of iterations (\textit{left}) as well as training duration (\textit{right}). ADGPS-4 and ADGPS-8 denote 4 and 8 pairs of local and global workers, respectively, while AGPS is an asynchronous run with a single pair of workers. Note that asynchronous training does slightly reduce the improvement per iteration, but substantially improves training time when multiple workers are used.}
  \label{fig:sim_learning_curves}
\end{figure}

\begin{figure}
  \centering
  \includegraphics[trim=0pt -7pt 0cm 0, clip=true,width=0.75\columnwidth]{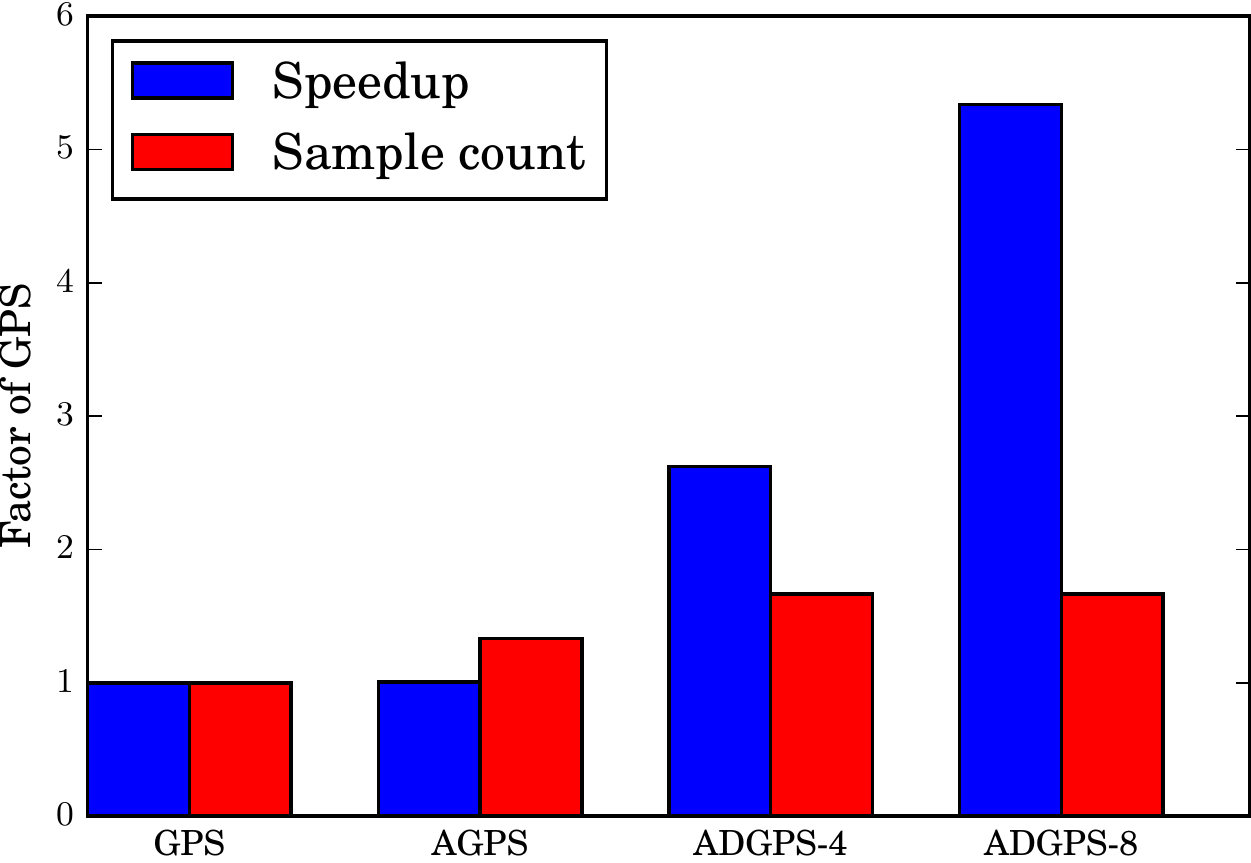}%
  \vspace{-5pt}
  \caption{Speedup in wall-clock training time and sample count comparison between GPS and the asynchronous variants, measured as the wall-clock time or sample count needed to reach a threshold cost value. Note that additional asynchronous workers incur only a modest cost in total sample count, while providing a substantial improvement in wall-clock training time.}
  \label{fig:sim_speedup}
\end{figure}

We ran guided policy search with and without asynchrony, and with increasing numbers of workers from 1 to 8. Figure~\ref{fig:sim_learning_curves} shows the average costs across four test instances for each setting of the algorithm, plotted against the number of trials and wall-clock time, respectively. ADGPS-4 and ADGPS-8 denote 4 and 8 pairs of local and global workers, respectively, while AGPS is an asynchronous run with a single pair of workers. Note that asynchronous training does slightly reduce the improvement in cost per iteration, since the local policies are updated against an older version of the global policy, and the global policy is trained on older data. However, the iterations themselves take less time, since the global policy training is parallelized with data collection and local policy updates. This substantially improves the learning rate in terms of wall clock time. This is illustrated in Figure~\ref{fig:sim_speedup}, which shows the relative improvement in wall-clock time (labeled as ``speedup'') compared to standard GPS, as well as the relative increase in sample complexity (labeled as ``sample count'') due to the slightly reduced policy improvement per iteration.

\subsection{Real-World Evaluation}
\label{sec:door_opening}

\begin{figure}
  \centering
  \includegraphics[trim=0 0 0 0, clip=true,width=\columnwidth]{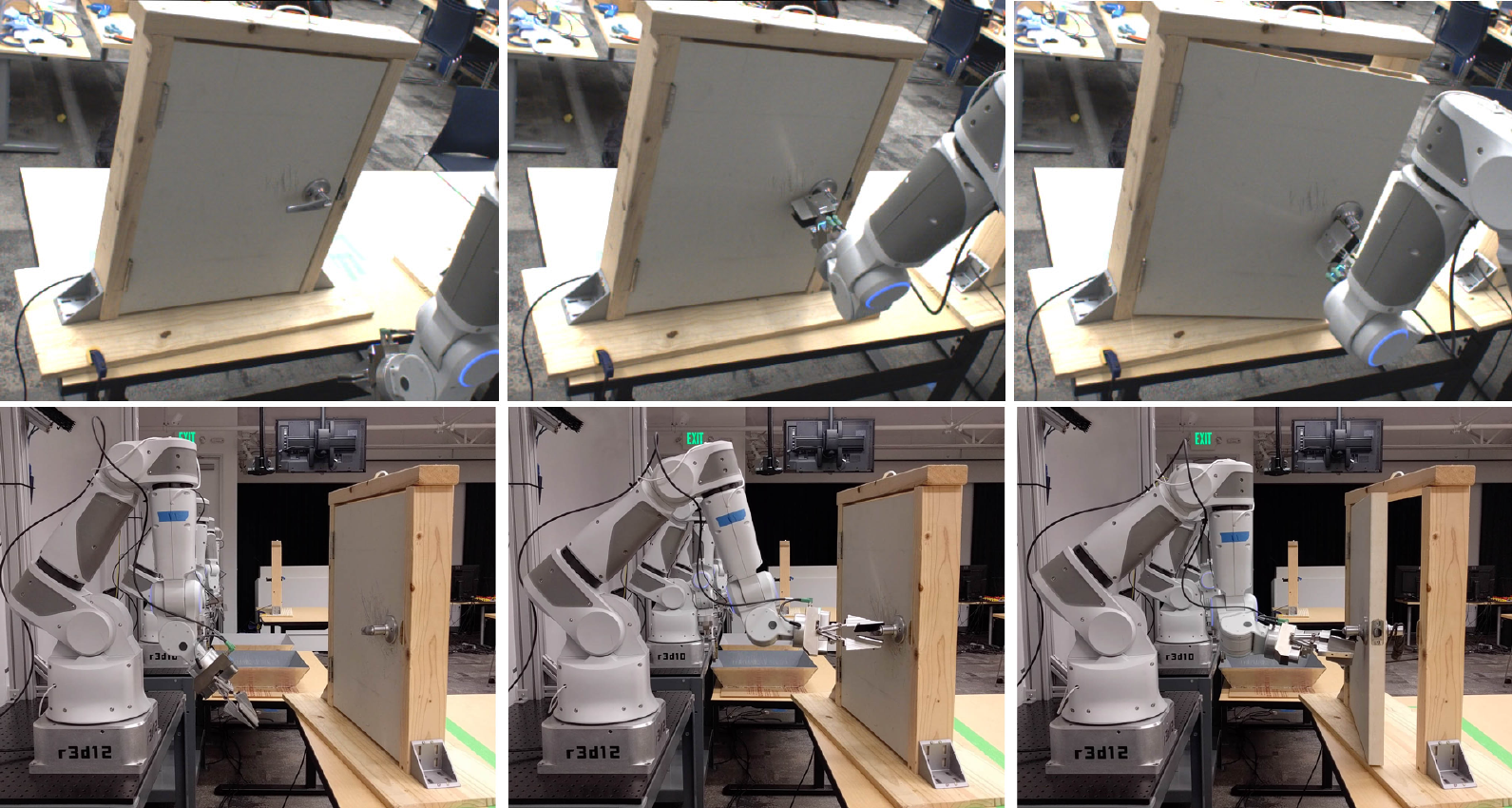}
  \vspace{-5pt}
  \caption{Door task execution. Top: sample robot RGB camera images used to control the robot. Bottom: side view of one of the robots opening a door.}
  \label{fig:door_opening_six}
\end{figure}

\begin{figure}
  \centering
  \includegraphics[trim=0 0 0 0, clip=true,width=\columnwidth]{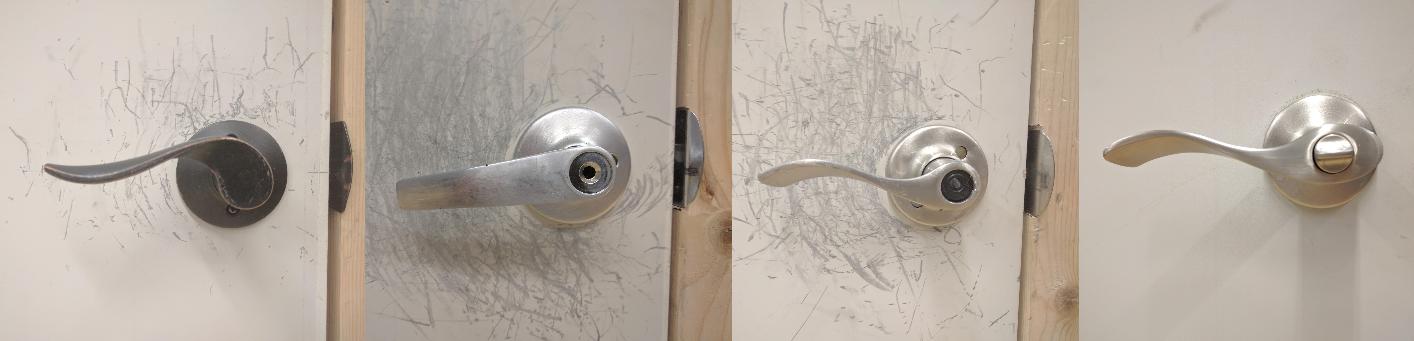}
  \vspace{-5pt}
  \caption{Variation in door handles used in the experiment described in Section~\ref{sec:door_opening}. The three handles on the left are used during training, and the handle on the right is used for evaluation.}
  \label{fig:door_handles}
\end{figure}

Our real-world evaluation is aimed at determining whether our distributed asynchronous system can effectively learn complex, nonlinear neural network policies, using visual inputs, and whether the resulting policies can generalize more effectively than policies learned on a single robot platform using the standard synchronous variant of GPS. To that end, we tackle a challenging real-world door opening task (Figure~\ref{fig:door_opening_six}), where the goal is to train a single visuomotor policy that can open a range of doors with visual and mechanical differences in the handle (Figure~\ref{fig:door_handles}), while also dealing with variations in the pose of the door with respect to the robot, variations in camera calibration, and mechanical variations between robots themselves.

We use four lightweight torque-controlled 7-DoF robotic arms, each of which is equipped with a two finger gripper, and a camera mounted behind the arm looking over the shoulder. The poses of these cameras are not precisely calibrated with respect to each robot. The input to the policy consists of monocular RGB images and the robot state vector as described in Section~\ref{sec:simulated_evaluation}. The robots are controlled at a frequency of 20Hz by directly sending torque commands to all seven joints. Each robot is assigned a specific door for policy training. The cost function is computed based on an IMU sensor attached to the door handle on the reverse side of the door. The desired IMU readings, which correspond to a successfully opened door, are recorded during kinesthetic teaching of the opening motion from human demonstration. We additionally add quadratic terms on joint velocities and commanded torques multiplied by a small constant to encourage smooth motions.

\begin{figure*}[t!]
\centering
\includegraphics[trim=0 0 0 0, clip=true,width=0.955\textwidth]{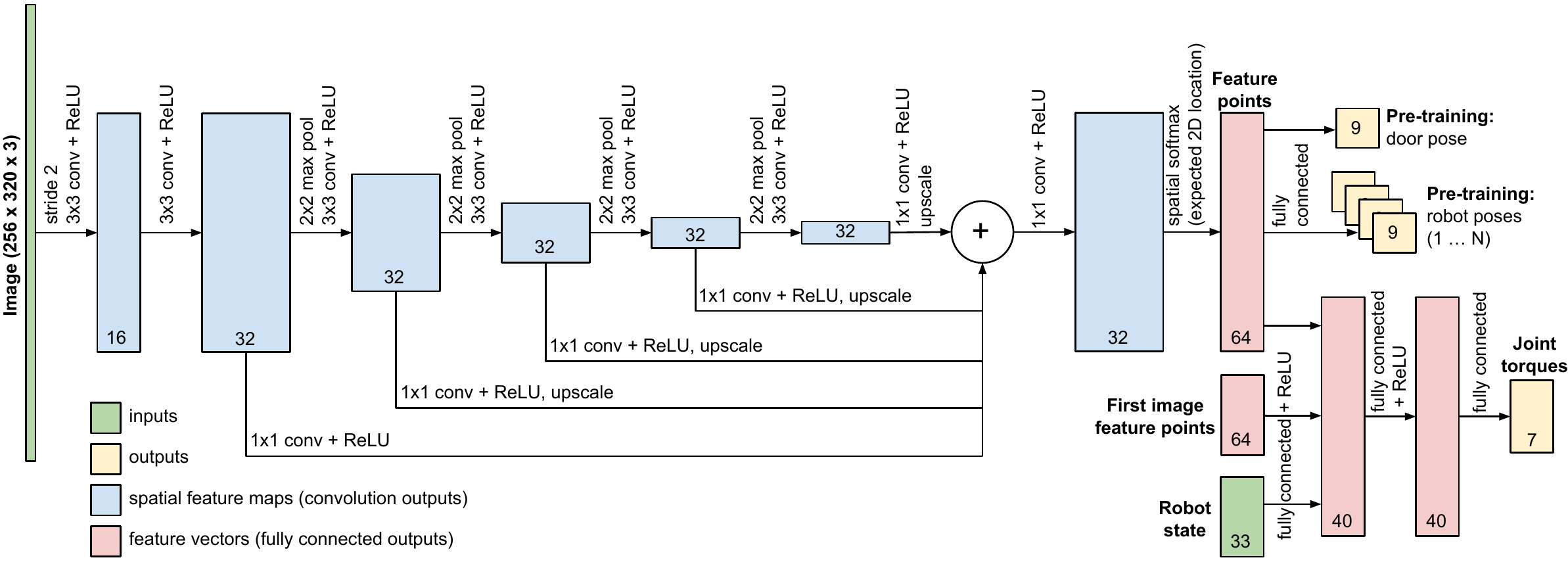}
\vspace{-6pt}
\caption{The architecture of our neural network policy. The input RGB image is passed through a 3x3 convolution with stride 2 to generate 16 features at a lower resolution. The next 5 layers are 3x3 convolutions followed by 2x2 max-pooling, each of which output 32 features at successively reduced resolutions and increased receptive field. The outputs of these 5 layers are recombined by passing each of them into a 1x1 convolution, converting them to a size of 125x157 by using nearest-neighbor upscaling, and summation (similar to~\cite{tompson2014joint}). A final 1x1 convolution is used to generate 32 feature maps. The spatial soft-argmax operator~\cite{Levine:2016} computes the expected 2D image coordinates of each feature. A fully connected layer is used to compute the object and robot poses from these expected 2D feature coordinates for pre-training the vision layers. The feature points for the current image are concatenated with feature points from the image at the first timestep as well as the 33-dimensional robot state vector, before being passed through two fully connected layers to produce the output joint torques.}
\label{fig:nn}
\vspace{-14pt}
\end{figure*}

The architecture of the neural network policy we use is shown in Figure~\ref{fig:nn}. Our architecture resembles prior work~\cite{Levine:2016}, with the visual features represented by feature points produced via a spatial softmax applied to the last convolutional response maps. Unlike in \cite{Levine:2016}, our convolutional network includes multiple stages of pooling and skip connections, which allows the visual features to incorporate information at various scales: low-level, high-resolution, local features as well as higher-level features with larger spatial context. This allows the network to generate high resolution features while limiting the amount of computation performed at high resolutions, enabling evaluation of this deep model at camera frame rates.

\subsubsection{Policy pre-training}

We train the above neural network policy in two stages. First, the convolutional layers are pretrained with a proxy pose detection objective. To create data for this pretraining phase, we collect camera images while manually moving each of the training doors into various poses, and automatically label each image by using a geometry-based pose estimator based on the point pair feature (PPF) algorithm~\cite{stefanppf}. We also collect images of each robot learning the task with PI$^2$ (without vision), and label these images with the pose of the robot end-effector obtained from forward kinematics. Each pose is represented as a 9-DoF vector, containing the positions of three points rigidly attached to the object (or robot), represented in the world frame. The door poses are all labeled in the camera frame, which allows us to pool this data across robots into a single dataset. However, since the robot endeffector poses are labeled in the base frame of each robot with an unknown camera offset, we cannot trivially train a single network to predict the pose label of any robot from the camera image alone. Hence, the pose of each robot is predicted using a separate output using a linear mapping from the feature points. This ensures that the 2-D image features learnt to predict the robot and door poses can be shared across all robots, while the 3-D robot pose predictions are allowed to vary across robots. The convolutional layers are trained using stochastic gradient descent (SGD) with momentum to predict the robot and door poses, using a standard Euclidean loss.

\subsubsection{Policy learning}

The local policy for each robot is initialized from its provided kinesthetic demonstration. We bootstrap the fully connected layers of the network by running four iterations of BADMM-based ADGPS with the PI$^2$ local policy optimizer. The pose of each door is kept fixed during bootstrapping. Next, we run 16 iterations of asynchronous distributed MDGPS with PI$^2$, where we randomly perturb each door pose at the start of every iteration. This sampling procedure allows us to train the global policy on a greater diversity of initial conditions, resulting in better generalization. The weights of the convolutional layers are kept frozen during all runs of GPS. In future work, it would be straightforward to also fine-tune the convolutional layers end-to-end with guided policy search as in prior work~\cite{Levine:2016}, but we found that we could obtain satisfactory performance without end-to-end training of the vision layers on this task.

\subsubsection{Results}

\begin{figure}
\centering
\begin{minipage}[c]{0.32\linewidth}
\begingroup
\fontsize{8pt}{12pt}\selectfont
\begin{tabular}{cc}
\toprule
Robot    & Success \\
         &  Rate \\
\midrule
1  & 90\%  \\
2  & 94\%  \\
3  & 90\%  \\
4  & 86\%  \\
\midrule
Mean & 90\% \\
\bottomrule
\end{tabular}
\endgroup
\par\vspace{0.7cm}
\end{minipage}\hfill
\begin{minipage}[c]{0.66\linewidth}
\includegraphics[trim=0 0 0 0, clip=true,width=\linewidth]{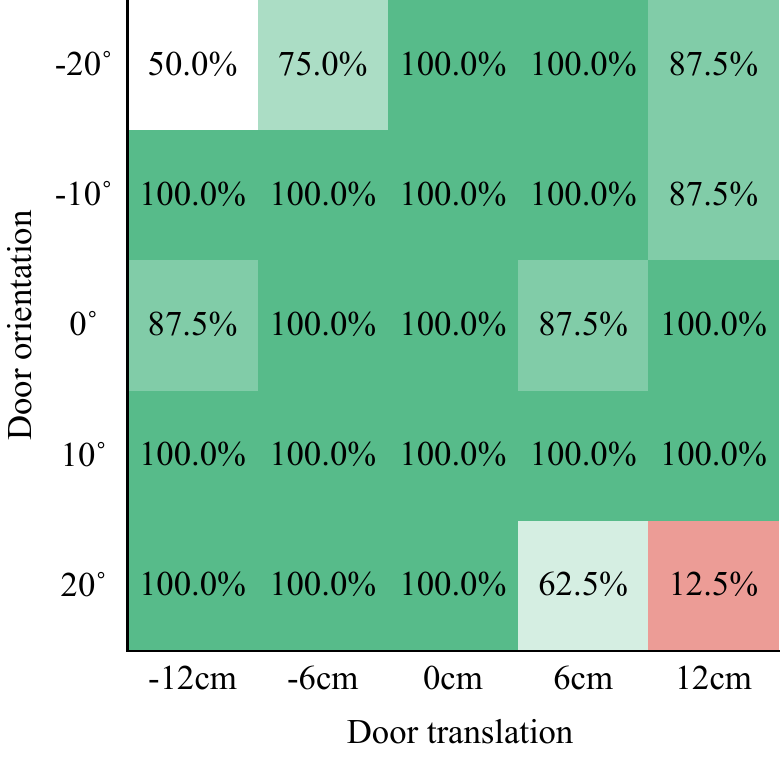}%
\par\vspace{0pt}
\end{minipage}
  \caption{Results from evaluating the visuomotor policy trained using ADGPS, using 50 trials per robot on a test door whose translations and orientations are sampled on a grid. \textit{Left}: Success rates per robot averaged over the sampling grid. \textit{Right}: Aggregate success rates across all robots for varying translations and orientations.}
  \label{fig:door_results}
\end{figure}

The trained policy was evaluated on a test door not seen during training time, by executing 50 trials per robot over a grid of translations and orientations. Figure~\ref{fig:door_results} shows the results obtained using the policy after training. We find that all four robots are able to open the test door in most configurations using a single global policy, showing generalization over appearance and mechanical properties of door handles, door position and orientation, camera calibration, and variations in robot dynamics. The lack of precise camera calibration per robot implies that the policy needs to visually track the pose of the door handle and the robot gripper, servoing it to the grasp pose. This is evident when watching the robot execute the learned policy (see video attachment): the initial motion of the robot brings the gripper into the view of the camera, after which the robot is able to translate and orient the gripper to grasp the handle before opening the door.

Furthermore, the trained policy was also evaluated with two test camera positions on the test door. The first camera position was arrived at by displacing the camera of one of the robots towards the ground by 5cm. The second position was arrived at by displacement that same camera away from the door by 4cm. The trained policy had success rates of 52\% and 54\% respectively with these two camera positions. In comparison, a successful policy that was trained on only a single robot and a single door using GPS with PI$^2$ as in~\cite{chebotar-icra2017} fails to generalize to either an unseen door or different camera positions.

\section{Discussion and Future Work}

We presented a system for distributed asynchronous policy learning across multiple robots that can collaborate to learn a single generalizable motor skill, represented by a deep neural network. Our method extends the guided policy search algorithm to the asynchronous setting, where maximal robot utilization is achieved by parallelizing policy training with experience collection. The robots continuously collect new experience and add it to a replay buffer that is used to train the global neural network policy. At the same time, each robot individually improves its local policy to succeed on its own particular instance of the task. Our simulated experiments demonstrate that this approach can reduce training times, while our real-world evaluation shows that a policy trained on multiple instances of different doors can improve the generalization capability of a vision-based door opening policy.

Our method also assumes that each robot can execute the same policy, which implicitly involves the assumption that the robots are physically similar or identical. An interesting future extension of our work is to handle the case where there is a systematic discrepancy between robotic platforms, necessitating a public and private component to each policy. In this case, the private components would be learned locally, while the public components would be trained using shared experience and pooled across robots. This could allow distributed asynchronous training to extend even to heterogeneous populations of robots, where highly dissimilar robots might share globally useful experience, such as the statistics of natural images, while robot-specific knowledge about, for example, the details of low-level motor control, would be shared only with physically similar platforms.

\section*{Acknowledgements}
We would like to thank Peter Pastor and Kurt Konolige for additional engineering, robot maintenance, and technical discussions, and Ryan Walker and Gary Vosters for designing custom hardware for this project.

\bibliographystyle{unsrtnat}
\renewcommand{\bibfont}{\footnotesize}
\bibliography{adgps}

\end{document}